\pdfoutput=1

\documentclass[11pt]{article}

\usepackage[preprint]{acl}

\usepackage{times}
\usepackage{latexsym}

\usepackage{booktabs}

\usepackage[T1]{fontenc}

\usepackage[utf8]{inputenc}

\usepackage{microtype}

\usepackage{inconsolata}

\usepackage{graphicx}

\usepackage{amsmath}

\title{Enhancing Dialogue Speech Recognition with \\ Robust Contextual Awareness via Noise Representation Learning}

\author{Wonjun Lee $^*$$^1$, San Kim $^*$$^2$ \and Gary Geunbae Lee $^{1,2}$\\
        $^1$ Department of Computer Science and Engineering, POSTECH, Republic of Korea \\
	 $^2$ Graduate School of Artificial Intelligence, POSTECH, Republic of Korea \\
    \{lee1jun, sankm, gblee\}@postech.ac.kr}

\begin{document}
\maketitle
\def\thefootnote{*}\footnotetext{Equally contributed}\def\thefootnote{\arabic{footnote}}
\begin{abstract}

Recent dialogue systems rely on turn-based spoken interactions, requiring accurate Automatic Speech Recognition (ASR). Errors in ASR can significantly impact downstream dialogue tasks. To address this, using dialogue context from user and agent interactions for transcribing subsequent utterances has been proposed. This method incorporates the transcription of the user's speech and the agent's response as model input, using the accumulated context generated by each turn.
However, this context is susceptible to ASR errors because it is generated by the ASR model in an auto-regressive fashion. Such noisy context can further degrade the benefits of context input, resulting in suboptimal ASR performance. In this paper, we introduce Context Noise Representation Learning (CNRL) to enhance robustness against noisy context, ultimately improving dialogue speech recognition accuracy.
To maximize the advantage of context awareness, our approach includes decoder pre-training using text-based dialogue data and noise representation learning for a context encoder. Based on the evaluation of speech dialogues, our method shows superior results compared to baselines. Furthermore, the strength of our approach is highlighted in noisy environments where user speech is barely audible due to real-world noise, relying on contextual information to transcribe the input accurately.

\end{abstract}

\section{Introduction}
Automatic Speech Recognition (ASR) is central in accurately interpreting human speech, serving as a fundamental resource for numerous subsequent downstream tasks. The advent of robust ASR modules, such as wav2vec2.0 \cite{baevski2020wav2vec} and Whisper \cite{radford2023robust}, has significantly enhanced the capabilities of ASR systems, facilitating their integration into a wide array of research and application domains. The integration of ASR modules into various works highlights the pivotal role of ASR in enhancing human-computer interaction, signifying a notable development in interactive technologies. 

Despite the successful advancement of the ASR system, its inaccuracy poses significant risks to the efficacy of downstream tasks, such as speech-to-text translation \cite{liu2020synchronous,le2024comsl,tang2021improving} and spoken language understanding \cite{serdyuk2018towards,arora2022espnet,huang2020learning}. These tasks predominantly rely on the textual output generated by ASR systems, highlighting the importance of accuracy in the initial speech recognition process. Especially for the dialogue system, the quality of the ASR system is paramount to ensure seamless interaction between user and dialogue agent, as models trained on written conversations perform poorly on spoken data \cite{kim2021robust}. To minimize the impact of ASR error on the dialogue model, various endeavors have been made. \citet{jiang-etal-2023-speech} used an ASR correction module which employs multiple ASR models, while others focused on augmenting data with plausible ASR errors \cite{park2023copyt5,wang2020data,tian2021todda}. However the limitation is evident as they primarily focus on the robustness of dialogue models, which may not address the core issue compared to directly rectifying ASR models.

Conversely, incorporating a context encoder for dialogue history to improve the ASR model has been proposed, resulting in notable performance enhancements \cite{ortiz2021bert,shenoy21_interspeech,hou2022bring,hori2020transformer}. 
Nevertheless, since the context is transcribed at each turn by the ASR model, it may contain errors, potentially disrupting the use of contextual information.

In this work, we present a novel Context Noise Representation Learning (CNRL) method to encode accurate contextual information, even from noisy ASR transcriptions. This approach aims to improve the performance of speech recognition in Task Oriented Dialogue (TOD) by minimizing the impact of ASR errors in dialogue history as context. 
Furthermore, we explore the advantages of decoder pre-training in context-aware ASR systems, emphasizing their improved robustness in noisy environments. The overall training pipeline can be decomposed by three steps: 1) Decoder pre-training on text-based dialogue data between user and agent. 2) ASR fine-tuning with speech encoder and context encoder jointly. 3) CNRL on context encoder to minimize the impact of ASR-noise context.
Our contributions are as follows:


\begin{itemize}
  \item We propose a novel training pipeline for dialogue speech recognition that leverages the dialogue history between user and agent.
  \item We demonstrate the effectiveness of CNRL by comparing it to various baseline models, showing a relative 13\% reduction in Word Error Rate (WER) compared to the current state-of-the-art ASR model \cite{radford2023robust}.
  \item In evaluations conducted in highly noisy environments, our model exhibits robust transcription accuracy, achieving up to a 31.4\% reduction in WER compared to the baseline.
\end{itemize}

\begin{figure*}[t!]
    \centering
    \includegraphics[width=0.9\linewidth]{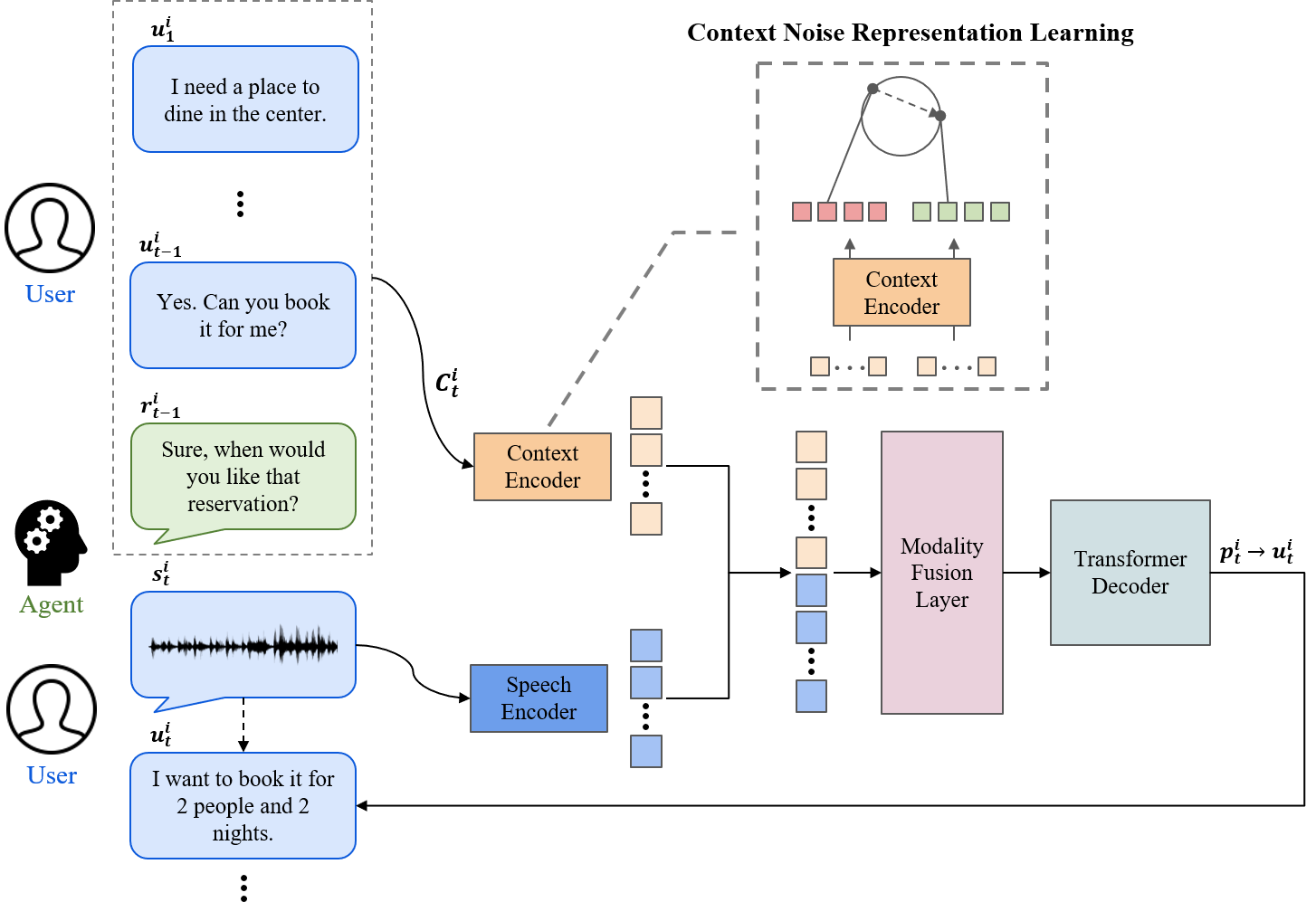}
    \caption{The architecture of a Context-Aware ASR(CA-ASR), featuring separate speech and context encoders to process the user's current speech $s^i_t$ and dialogue history $C^i_t$, respectively. These representations are concatenated and fused using a modality fusion layer and transcribed to the predicted user utterance $p^i_t$ by the transformer decoder. The predicted user utterance will be added to context ($p^i_t \xrightarrow{} u^i_t$) for the next turn ($t+1$). After the training, the context encoder can improve itself by our CNRL method, detailed in Figure \ref{fig:cnrl} and Section \ref{sec:method.CNRL}.
    }
    \label{fig:figure1}
\end{figure*}

\section{Related work}


\subsection{Context-aware speech recognition}
Several studies have shown that leveraging contextual information in dialogue scenarios can enhance ASR performance. 
\citet{shenoy21_interspeech} used a context carry-over mechanism to enhance the recurrent model's accuracy. \citet{hou2022bring} proposed utilizing a context encoder in RNN-T architecture, adopting the semantic embedding of dialogue context from BERT \cite{bert}. \citet{hori2020transformer} targeted considering long-context by sliding-window fashion. \citet{wang2023speech} and \citet{wang2024retrieval} proposed an audio-augmented retriever to directly transcribe and track the dialogue state. These Context-Aware ASR(CA-ASR) models have a potential drawback: the context generated for each turn is based on ASR transcriptions, which inevitably contain errors, potentially degrading context-awareness. In this paper, we introduce the CNRL method, which trains only the context encoder independently. The goal is to enable the context encoder to produce similar encoding for noisy (ASR output) contexts to match clean context.

\subsection{Decoder pre-training}

Compared to pre-training encoder layers \cite{baevski2020wav2vec, hsu2021hubert,chen2022wavlm}, pre-training the decoder for ASR has received comparatively less attention. Notably, in scenarios where input speech is flawed or incomplete, the decoder can still play a crucial role in transcribing user utterances by leveraging contextual language modeling. To harness the decoder's capabilities, the use of external datasets like phoneme-to-grapheme paired data \cite{masumura2020phoneme} or text data \cite{gao2021pre} has been suggested. This approach enables the model to benefit from numerous external, non-paired data sources. \citet{tsunoo2023decoder} trained decoder for both ASR task and language modeling task, enabling improved linguistic understanding and leading to better ASR performance. Following these works, we pre-trained the decoder for a context-aware ASR model using voluminous text-only data. Specifically, we focus on turn-based dialogue data between user and agent, where each utterance is highly related to each other. 

\subsection{Noise Representation Learning}

Noise in input data is inevitable in various forms across many datasets. Training models with such data negatively impacts their generalization performance. To address this challenge, numerous studies have adopted contrastive learning to enhance model robustness. \citet{ma2023noise} improved named entity recognition performance by employing a token-level dynamic loss function and contrastive learning, leveraging noisy data and accounting for noise-distribution changes during training. \citet{xu2023simcse} enhanced contrastive learning through a dimension-wise method to mitigate feature corruption in sentence embeddings. \citet{sun2023noise} used a K-NN graph to identify confident samples and applied mixup supervised contrastive learning to create robust representations, leading to improved relation extraction performance. \citet{zheng2023robust} utilized both class-wise and instance-wise contrastive learning in their novel representation learning module. 
In this work, we adopt representation learning to enhance context awareness when noisy ASR transcriptions are used for context. The proposed CNRL is integrated solely with the context encoder in the CA-ASR model to minimize training costs.

\section{Methodology}

\subsection{Preliminary}

We define $D^{i}_t$ as the turn-based dialogue dataset for turn $t$ in the $i$-th dialogue, which includes the speech input $s^i_t$, the corresponding text labels $u^i_t$ (transcriptions) of user utterances, and the turn-based dialogue history $C^i_t = (u^i_1, r^i_1, ..., u^i_{t-1}, r^i_{t-1})$, accumulating up to turn $t-1$, where $r^i_t$ represents the agent's response at turn $t$. Each dialogue instance at the $k$-th turn, denoted as ($u^i_k$, $r^i_k$), comprises a single-turn conversation consisting of both a user utterance and an agent response. 
During inference, the predicted utterance (transcription) from model $p^i_t$ is used instead of $u^i_t$ for user utterance to form context $C^i_t$.





The CA-ASR model integrates the user's speech and dialogue history. For each turn $t$, the model predicts the current user utterance $u^i_t$ from the speech input $s^i_t$ and the context $C^i_t$. The dialogue history comprises text logs from both the user and the agent, where the user's speech is transcribed in real-time, while the agent's responses are given in text format. 
To transcribe the user's speech at turn $t$, the model draws upon past conversations from turn $1$ to $t-1$. Utilizing an encoder-decoder architecture for the CA-ASR model, dedicated encoders initially process each input type—speech and text. These encodings are then concatenated and fused through a modality fusion layer, yielding a fused representation. Subsequently, the fused representation is passed through a decoder layer to transcribe the user utterance. Figure \ref{fig:figure1} illustrates the CA-ASR architecture, highlighting the interaction between user utterances and agent responses.

\subsection{Decoder pre-training for Dialogue}
\label{sec:pre}
We adopt a pre-training method specifically targeting decoders in the CA-ASR model. This method employs an encoder-decoder architecture, where the model takes the text-form dialogue history $C^i_t$ as input. For the output, since the decoder is eventually used for transcribing user utterances, it aims to predict the next user utterance $u^i_t$. Additionally, the utilization of text data as input enables the training process to use external text datasets, further enhancing the decoder's performance. We demonstrate this efficacy in Section \ref{sec:5.2} . This approach enables the decoder to anticipate the subsequent user utterance based on contextual information derived from the dialogue history. 
This training method is particularly effective because dialogues in TOD are more predictable from the dialogue history than other types of conversations. In typical user-agent interactions, the agent often asks specific questions, and the user responds with relevant answers, making the dialogue structure more consistent and easier to predict.

When integrated into the CA-ASR model and fine-tuned for ASR tasks, the pre-trained decoder can significantly enhance transcription performance. By leveraging its ability to anticipate user responses from the agent's response (or the entire dialogue history), the decoder contributes to more accurate and robust transcription results, even with imperfect input speech, such as noisy audio signals.



\begin{figure}[t]
    \centering
    \includegraphics[width=1\linewidth]{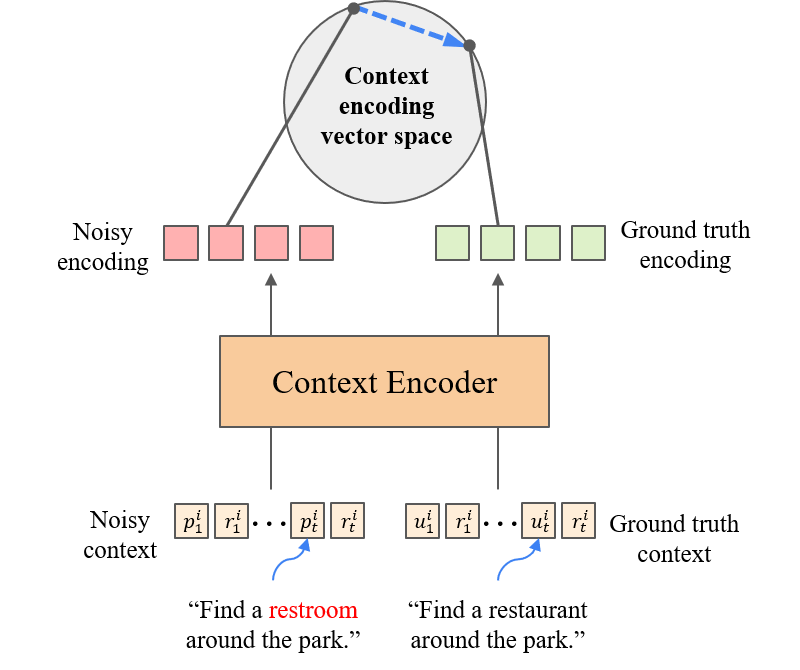}
    \caption{\textbf{Context Noise Representation Learning:} The noisy context including user utterances generated by the CA-ASR model during inference ($p_t^{i}$), and the ground truth context with clean user utterances ($u_t^{i}$), are encoded by the context encoder. The noisy encoding is adjusted to closely match the ground truth encoding in the context encoding vector space.}
    \label{fig:cnrl}
\end{figure}




\subsection{Context Noise Representation Learning}
\label{sec:method.CNRL}

During inference, the CA-ASR model uses context from previous transcriptions of user utterances and agent responses. However, inaccuracies in the ASR-generated transcriptions can degrade the advantage of using context, as training typically uses only ground truth context for each turn. To address this, we introduce CNRL. This method involves an additional training step where the model transcribes and utilizes noisy transcriptions to train the context encoder in a representation learning manner, as illustrated in Figure \ref{fig:cnrl}. The context encoder is fine-tuned to generate similar encoding for noisy input context as it does for the ground truth context. This method focuses solely on enhancing the context encoder, maintaining training efficiency.

To create the training set for CNRL, we first generate noisy transcriptions using the CA-ASR model with the ASR training set (See Section \ref{sec:dataset}) divided into 10 folds. In each fold, 90\% of the training set is used to train the CA-ASR model, and the remaining 10\% is used to generate noisy ASR transcriptions. By iterating through all 10 folds, we obtain a complete noisy context training set.
The dataset for CNRL comprises pairs of noisy and ground truth contexts, each containing multiple conversation turns. Each turn pairs a user utterance with an agent response, except for the initial turn, which consists only of the user's utterance.

We trained context encoder with cosine embedding loss:

\small
\begin{equation}
\text{loss}(x,y) = \begin{cases}
1 - \cos(x_1, x_2), & \text{if } y = 1 \\
\max(0, \cos(x_1, x_2) - \text{margin}), & \text{if } y = -1
\end{cases}
\end{equation}
\normalsize

\noindent Where $x_1$ is the encoding vector from the context encoder within the ASR-generated context and $x_2$ is the encoding from ground truth context. $y$ is the label that indicates these two ($x_1$ and $x_2$) are of the same class($y=1$) or not($y=-1$). Since we trained the context encoder to generate a similar output encoding for the noisy input ($x_1$) to match the clean ground truth ($x_2$), we set $y = 1$ for training. During training, $x_1$ gets close to $x_2$ on context encoding vector space, ensuring the context encoder produces similar encoding for a given noisy context. By using CNRL, the context encoder can maintain accurate context information, leading to improved speech recognition accuracy.

\section{Experimental setup}


\subsection{Datasets}
\label{sec:dataset}

\noindent\textbf{The DSTC11 Challenge Dataset} The DSTC11  \cite{dstc11} dataset is derived from the MultiWoZ 2.1 \cite{eric-etal-2020-multiwoz} by adding speech recordings and synthesized voices generated by a TTS model. The training set is built using the TTS model, while the evaluation sets are recorded by human volunteers. 
Each dialogue consists of audio files of user utterances paired with corresponding agent responses. In every dialogue, the user initiates the conversation, making the first user utterance has no preceding context.

Since the official transcription for the DSTC11 test split (test-dstc11.human-verbatim) is not publicly available, we evaluate our experiments on the DSTC11 development split with human recording (dev-dstc11.human-verbatim)\footnote{https://storage.googleapis.com/gresearch/\newline dstc11/dstc11\_20221102a.html} as test set. Additionally we randomly sampled 3000 audios from the training set and used them as our development set during training.

The DSTC11 training set consists of 8,434 dialogues comprising 56,750 user utterances synthesized by four TTS voices, generating a total of 227,000 audio files. Our development set, randomly sampled from the training set, contains 3000 user utterances and is excluded from the training data. The test set includes human recordings of 7,374 user utterances from 1,000 dialogues. The average audio duration is 3.31 seconds for the training and development sets and 5.35 seconds for the test set.


\noindent\textbf{Evaluation in Noisy Environments} Environmental noise is a significant challenge for ASR systems in real-world scenarios. However, contextual information can mitigate this issue. To test our ASR system's resilience to real-world noises, we use the ESC-50 dataset \cite{ESC-50}, which includes 50 classes of common urban noises, such as drilling and sirens. Noise samples are randomly selected from 2000 audio files and injected into our test set at Signal-to-Noise Ratios (SNR) of 20dB and 0dB, representing soft and hard noise conditions, respectively. This evaluation replicates challenging acoustic environments to test the ASR system's robustness rigorously. Note that the noisy audio is used exclusively for evaluation, not training. Our goal is to show that contextual information can be helpful in noisy environments where the audio signal is significantly degraded.




\noindent\textbf{Decoder pre-training} 
To facilitate the use of context information, we first trained CA-ASR's decoder using exclusively text-based data before ASR fine-tuning. For this purpose, we employ large datasets of turn-based dialogue text, combining the Schema-Guided Dialogue (SGD) \cite{rastogi2020towards}  dataset with the DSTC11 text dataset to pre-train the decoder. SGD consists of over 20,000 task-oriented conversations between human and virtual assistant. From 8434 English dialogues from DSTC11 and approximately 16,000 English dialogues from the SGD training dataset, we use about 260,000 turn conversations. To evaluate the effect of decoder pre-training, we varied the volume of text data used for this process.  
The effects of these variations are detailed in Table \ref{tab:decoder-compare}.

\subsection{Model configuration}

\noindent\textbf{Baselines} We compare our CA-ASR model against several baselines, including those reported in DSTC11 \cite{dstc11} and the current state-of-the-art ASR model Whisper \cite{radford2023robust}. Additionally, we present a model that uses wav2vec2.0 \cite{baevski2020wav2vec} as the encoder and BART \cite{bart} as the decoder. This model shares the same architecture as the CA-ASR model, except for removing the context encoder and modality fusion. For transcription post-processing, we normalize common English patterns (e.g., "I've" to "I have"), remove punctuation, and normalize digits to ensure a fair comparison between models.

\noindent\textbf{Context-Aware ASR} Compared to the baselines, the CA-ASR model leverages previous user utterances and agent responses as textual input to enhance transcription accuracy. To encode this contextual information, CA-ASR uses the BART encoder as the context encoder. The speech encoder is wav2vec2.0 with the checkpoint \textit{wav2vec2-large-960h}\footnote{https://huggingface.co/facebook/wav2vec2-large-960h}, and the pretrained BART encoder and decoder with the checkpoint \textit{bart-large}\footnote{https://huggingface.co/facebook/bart-large} are utilized as the context encoder and the CA-ASR decoder, respectively. 
Given that the maximum token length for BART-large is limited to 1024, we truncate the context to the last 1024 tokens if necessary.

For modality fusion, the wav2vec2.0 speech encoder and the BART context encoder each produce hidden representations with dimensions of $\text{token} \times 1024$. Since the BART decoder requires an encoder hidden state with a dimension of 1024, we concatenate these hidden representations along the 1024 dimension. This concatenated representation is then passed through a linear layer (1024, 1024) with ReLU activation to create a fused representation. This fused representation is subsequently fed into the BART decoder to transcribe the user utterance.

Total parameter size of our model is 774M, consisting of 315M for the speech encoder, 203M for the BART context encoder, 254M for the BART decoder, and 1M for the linear fusion layer.

\subsection{Training configuration}
\label{sec:training}

Our training pipeline consists of three sequential steps: decoder pretraining, ASR fine-tuning with audio masking, and CNRL. We evaluate the effect of each step in the subsequent Result \& Analysis section.

\noindent\textbf{Decoder pre-training} We initially adopt the BART encoder-decoder model to pre-train the decoder, which is subsequently used for ASR fine-tuning. The optimization is performed using the AdamW algorithm \cite{loshchilov2017decoupled} with $(\beta_1, \beta_2)=(0.9, 0.999)$, learning rate of 5e-5, weight decay of 1e-5, and a batch size of 32. We select the best model based on the lowest validation loss over 10 epochs of training, spanning 50 hours. The encoder functions as the context encoder, while the decoder serves as the transformer decoder in the CA-ASR model. Utilizing Cross-Entropy loss, we aim to input the dialogue history with the agent's response, which is the last turn, into the encoder and generate the user's response as the output from the decoder.

\noindent\textbf{ASR fine-tuning} In ASR fine-tuning stage, a speech encoder (wav2vec2.0) is attached to the pre-trained BART decoder from decoder pre-training. We adopt a batch size of 64 and an Adam optimizer with a learning rate of 2e-5. Across 10 epochs of training for 20 hours, the model with the lowest WER on development set at the end of each epoch was chosen as the best model for the speech encoder.

\noindent\textbf{Audio masking}
\label{sec:audio_masking}
Motivated by other multi-modal ASR study \cite{shi2022avhubert}, a small portion of the speech data is obscured by masking to reduce the model's reliance on speech input. Specifically, 10\% of speech data are randomly chosen for masking, and each selected data is masked for 20\% of its total duration. Note that this configuration of masking probability and duration was empirically determined to yield optimal results in our experiments, with the proportion of masked data and masking length varied between 10\% to 30\% and 10\% to 50\%, respectively. To implement the masking process, we segment each audio into discrete chunks of 1-second duration. These chunks serve as the minimum unit for the masking, e.g. in an audio input with a duration of 10 seconds, two randomly chosen chunks would be masked. Unless otherwise specified, all results of the CA-ASR model include audio masking during training.

\noindent\textbf{CNRL Setup} We utilized the noisy context training set from the 10-folds described in Section \ref{sec:method.CNRL}. The average WER for the noisy context was 6.53\% across the 10 folds. We filtered out transcriptions with a WER exceeding 20\% to prevent interference with CNRL, resulting in the exclusion of 8.2\% of the noisy context training set.
We evaluated the effect of CNRL noisy context data by modifying the dialogue turns and introducing arbitrary word drops. For arbitrary word drop, we remove words for user utterances from golden context by 10\% of change for each word and iterate it until we match the WER for each dialogue up to 6.5\%, which is similar to WER with 10-folds.
The training data setups are listed below:

\begin{itemize}
\item \textbf{S1}: Arbitrarily remove words from the golden context (user utterance only) to match an average WER of 6.5\%.
\item \textbf{S2}: Using the 10-fold training set, only the last user utterance contains noisy text.
\item \textbf{S3}: Using the 10-fold training set, all user utterances may contain noisy text.
\item \textbf{S4}: Combining S1 with S3. If a user utterance for each turn does not contain noisy text, arbitrary word drops are applied to increase the noise. 
\end{itemize}
Unless otherwise specified, subsequent experimental results with CNRL use the \textbf{S4} setup.
We use a batch size of 128 and the Adam optimizer with a learning rate of 5e-4. Training is conducted for up to 5 epochs, selecting the epoch with the lowest cosine embedding loss on our development set.

All experiments are conducted using 4 NVIDIA A6000 GPUs.

\begin{table*}[]
\resizebox{\textwidth}{!}{%
\begin{tabular}{lll|rrr}
\toprule
\multicolumn{3}{c|}{\textbf{Configurations}}                                 & \multicolumn{3}{c}{\textbf{Audio Noise Level}}                                                                       \\ \midrule
\multicolumn{1}{c}{\textbf{Model}}  & \multicolumn{1}{c}{\textbf{Modality}} & \multicolumn{1}{c|}{\textbf{Parameter size}} & \multicolumn{1}{c}{\textbf{No Noise}} & \multicolumn{1}{c}{\textbf{SNR:20dB}} & \multicolumn{1}{c}{\textbf{SNR:0dB}} \\ \hline \midrule
DSTC11 RNN-T \citep{dstc11}                       & Speech                                      & 220M & 11.90\%                               & \multicolumn{1}{l}{-}                 & \multicolumn{1}{l}{-}                \\
DSTC11 Whisper \citep{dstc11}*                    & Speech                                      & 1550M & 8.50\%                                & \multicolumn{1}{l}{-}                 & \multicolumn{1}{l}{-}                \\
Whisper-large-v2 \citep{radford2023robust}**                 & Speech                                      & 1550M & 8.10\%                                & 8.45\%                                & 14.82\%                              \\ \midrule
Wav2Vec2.0+BART (baseline)          & Speech                                      & 569M & 9.23\%                                & 11.89\%                               & 18.45\%                              \\ \midrule
CA-ASR (Ours)               & Speech+Text                                    & 774M & 7.92\%                                & 8.23\%                                & 15.65\%                              \\
    +CNRL                 & Speech+Text                                    & 774M & 7.66\%                                & 8.10\%                                & 15.03\%                              \\
    +Decoder Ptr.           & Speech+Text                                    & 774M & 7.39\%                                & 7.51\%                                & 13.33\%                              \\
    +Decoder Ptr. \& CNRL  & Speech+Text                                    & 774M & \textbf{7.04\%}                       & \textbf{7.24\%}                       & \textbf{12.65\%}                     \\ \bottomrule
\end{tabular}%
}
\caption{WER comparison of various baselines and CA-ASR settings under different noise conditions. Our proposed CA-ASR model is evaluated with and without Context Noise Representation Learning (CNRL) and Decoder Pretraining (Decoder Ptr.) enhancements. * : reported. **: re-evaluated with our post-processing.}
\label{tab:main}
\end{table*}

\section{Result \& Analysis}

\subsection{Context Aware-ASR}
Table \ref{tab:main} illustrates the WER across various models and noise levels. The CA-ASR model significantly improves performance on our test set, reducing relative WER by \textbf{33.4\%} compared to the RNN-T \cite{dstc11} baseline \textbf{(7.92\% vs. 11.90\%)} and by \textbf{14.2\%} compared to the wav2vec2.0 with BART baseline, even without additional methods like CNRL or decoder pre-training. This highlights the advantage of using multi-modality with a context encoder for dialogue speech recognition.

Decoder pre-training further enhances the performance of the CA-ASR model, significantly reducing relative WER by \textbf{6.7\%}, especially under severe noise conditions (SNR:0dB) where the voice is barely audible. This is expected since the decoder is initially tuned to the dialogue domain, enabling it to predict the user's subsequent probable response from the context even with incomplete speech input.

The benefits are maximized when CNRL is applied, resulting in a relative WER reduction of \textbf{11.1\%} in clean conditions and \textbf{19.1\%} in noisy environments compared to the basic CA-ASR model. Since CNRL is designed to make the context encoder resilient to context errors, it significantly enhances the model's robustness against strong noise.

Under the noisy audio test set (refer to Section \ref{sec:dataset}), each model's performance declines as the noise level increases (SNR:20dB to SNR:0dB). However, incorporating decoder pre-training and CNRL significantly mitigates this performance drop compared to the basic CA-ASR model (\textbf{12.65\% vs. 15.65\%}).

While the Whisper model shows robust performance under severe noise conditions (SNR:0dB), our CA-ASR model with CNRL and decoder pre-training demonstrates even greater robustness (\textbf{12.65\% vs. 14.82\%}).

\begin{table}[t]
\resizebox{\columnwidth}{!}{%
\begin{tabular}{lllr}
\toprule
\multicolumn{1}{c}{\textbf{Model}} & \multicolumn{1}{c}{\textbf{Input Dialogue}} & \multicolumn{1}{c}{\textbf{Decoder Pre-traing}} & \multicolumn{1}{c}{\textbf{WER}} \\ \midrule \midrule
baseline                           & -                                                & BART\citep{bart}             & 9.23\%                           \\
baseline                           & -                                                & + MultiWoZ 2.1                                 & 8.95\%                           \\
baseline                           & -                                                & + SGD                                          & 8.88\%                           \\ \midrule
CA-ASR                             & single-turn                                           & BART                                           & 8.14\%                           \\
CA-ASR                             & single-turn                                          & + MultiWoZ 2.1                                 & 7.98\%                           \\
CA-ASR                             & single-turn                                          & + SGD                                          & 7.64\%                           \\ \midrule
CA-ASR                             & multi-turn                                           & BART                                           & 7.92\%                           \\
CA-ASR                             & multi-turn                                           & + MultiWoZ 2.1                                 & 7.45\%                           \\
CA-ASR                    & multi-turn                                           & + SGD                                          & \textbf{7.39\%}                  \\ \bottomrule
\end{tabular}%
}
\caption{WER across various accumulated datasets and a number of turn-takings. Note that CNRL and noise evaluation are not applied in this result to focus on the efficacy of decoder pre-training.}
\label{tab:decoder-compare}
\end{table}

\subsection{Decoder Pre-training for Dialogue}
\label{sec:5.2}
Table \ref{tab:decoder-compare} demonstrates the effectiveness of pre-training the decoder with varying the number of turns and pre-training dataset sizes. Note that the baseline model is the same as in Table \ref{tab:main}, consisting only of a speech encoder (wav2vec2.0) and a BART decoder. As illustrated, pre-training the decoder on the dialogue domain benefits both the speech-only model (baseline) and the speech-text multimodal model (CA-ASR). Compared to the best result of baseline, the inclusion of the context encoder leads to significant improvements, resulting in a relative WER reduction of approximately \textbf{16.7\%} at best in CA-ASR with multi-turn (\textbf{8.88\% vs. 7.39\%}). This finding suggests that the efficacy of pre-training the decoder is maximized when the model incorporates information from previous dialogues. Additionally, the WER of CA-ASR with multi-turn improves relatively by up to 6.7\% as the dataset size increases (adding SGD), indicating the utility of incorporating external datasets as long as they involve user-agent conversations. Moreover, models considering multiple turns of dialogue exhibit a relatively 3.2\% better WER compared to those considering a single turn, as shown in the comparison of best results (\textbf{7.64\%} vs \textbf{7.39\%}). This highlights the importance of considering a longer context.


\subsection{Effect of Audio masking}

Since audio masking can serve as data augmentation, we conducted additional experiments to compare the performance improvement between the baseline (speech-only) model and the CA-ASR (multimodal) model. As shown in Table \ref{tab:audio-masking}, audio masking enhances ASR performance in both the baseline and CA-ASR models. While the baseline models exhibit marginal performance improvements of about 0.6\% in clean sample evaluations, CA-ASR benefits from audio masking with a \textbf{5.5\%} relative WER reduction. The improvement in CA-ASR becomes more pronounced in noisy environments as noise levels increase. Although the WER is highest at SNR:0dB, indicating the strongest noise, the relative WER reduction is \textbf{11.4\%}, compared to 8.12\% at SNR:20dB. These results suggest that while audio masking is beneficial in both clean and noisy environments, its effect is maximized when the model can utilize contextual information.

\begin{table}[t]
\resizebox{\columnwidth}{!}{%
\begin{tabular}{llrrr}
\toprule
\textbf{Model (Modality)} & \textbf{Audio Masking} & \multicolumn{1}{c}{\textbf{No Noise}} & \multicolumn{1}{c}{\textbf{SNR:20db}} & \multicolumn{1}{c}{\textbf{SNR:0db}} \\ \midrule\midrule
baseline (Speech)         & No                     & 8.94\%                                & 11.20\%                               & 18.02\%                              \\
baseline (Speech)         & Yes                    & 8.88\%                                & 10.58\%                               & 17.61\%                              \\ \midrule
CA-ASR (Speech + Text)    & No                     & 7.45\%                                & 7.88\%                                & 14.28\%                              \\
CA-ASR (Speech + Text)    & Yes                    & \textbf{7.04\%}                       & \textbf{7.24\%}                       & \textbf{12.65\%}                     \\ \bottomrule
\end{tabular}%
}
\caption{WER comparison between madality and audio masking in clean and noisy samples. Each model's decoder is pre-trained with Multio-WoZ 2.1 and SGD, and CNRL is additionally applied to CA-ASR.}
\label{tab:audio-masking}
\end{table}



\subsection{Context Noise Representation Learning}

To investigate the impact of noise data on CNRL, we conducted experiments using different types of noise (S1-S4) as described in the CNRL setup in Section \ref{sec:training}. 
In Table \ref{tab:result.cnrl}, compared to the model without CNRL, S1 (which arbitrarily removes words) degraded performance, indicating that using only artificial noise is not beneficial for CNRL. S2 and S3, which use real ASR noise from 10-fold data generation, showed better performance, with multi-turn noise (S3) outperforming single-turn noise (S2).

In our evaluation, we found that S4, which combines S1 with S3, performed the best, with WERs of 7.04\%, 7.24\%, and 12.65\% for No-Noise, SNR:20dB, and SNR:0dB conditions, respectively. For comparison, we evaluated our model with ground truth context during inference without CNRL, serving as the upper bound of our experiment. As expected, using ground truth context showed robust results across noise levels, while CNRL with S4 produced similar results with a small margin. This demonstrates that CNRL enables the context encoder to handle noisy contexts effectively, generating representations close to the ground truth.

We also experimented with training the full CA-ASR model, not just the context encoder, using S4 with corresponding audio for ASR fine-tuning. Training the full model showed lower performance gains than CNRL (\textbf{7.04\% vs. 7.24\%}) and required much larger training costs. We believe this is because training all components with noisy data can disrupt optimization. CNRL allows us to maintain ASR performance against noisy contexts while keeping training efficient.

\begin{table}[t]
\resizebox{\columnwidth}{!}{%
\begin{tabular}{lrrr}
\hline
\textbf{CNRL.}      & \multicolumn{1}{l}{\textbf{No Noise}} & \multicolumn{1}{l}{\textbf{SNR:20db}} & \multicolumn{1}{l}{\textbf{SNR:0db}} \\ \hline\hline 
No                  & 7.39\%                                & 7.51\%                                & 13.33\%                              \\ \hline
S1                  & 7.53\%                                & 7.45\%                                & 13.45\%                              \\
S2                  & 7.30\%                                & 7.41\%                                & 12.94\%                              \\
S3                  & 7.22\%                                & 7.29\%                                & 12.83\%                              \\
S4                  & 7.04\%                                & \textbf{7.24\%}                       & 12.65\%                              \\ \hline
Ground Truth Context*     & \textbf{7.01\%}                       & 7.25\%                                & \textbf{12.28\%}                     \\ \hline
full fine-tune w/ S4 & 7.24\%                                & 7.63\%                                & 13.50\%                              \\ \hline

\end{tabular}%
}
\caption{CNRL result on different training data settings (S1, S2, S3 and S4) including evaluation result with ground truth context (*) and full fine-tuning result.}
\label{tab:result.cnrl}
\end{table}

\section{Conclusion}
This study introduced Context Noise Representation Learning (CNRL) to improve context-aware ASR systems, especially in noisy environments. By integrating decoder pre-training with dialogue data, ASR fine-tuning, and CNRL, we significantly reduced transcription errors.
Our training pipeline demonstrated significant improvements in dialogue speech recognition, even in noisy environments where speech input is defective. Experiments showed CNRL's efficacy, reducing Word Error Rate (WER) by up to 11.1\% in clean conditions and 19.1\% in noisy settings. By making the model more robust against noisy context, our approach consistently outperformed baselines in various settings.

\newpage
\section*{Limitations}
Due to the scarcity of spoken turn-based dialogue datasets, we could only validate our method on a single dataset DSTC11. However validating on the various test datasets would improve its credibility if applicable.

Our primary goal is to enhance ASR performance. However, these enhancements could be even more valuable for downstream Dialogue State Tracking (DST) tasks. Future work could explore optimizing ASR specifically for DST applications to further increase the impact and value of our contributions.

\section*{Acknowledgement}
This work was supported by the Technology Innovation Program(20015007, Development of Digital Therapeutics of Cognitive Behavioral Therapy for treating Panic Disorder) funded By the Ministry of Trade, Industry \& Energy(MOTIE, Korea).

\noindent This research was supported by the MSIT(Ministry of Science and ICT), Korea, under the ITRC(Information Technology Research Center) support program(IITP-2024-2020-0-01789) supervised by the IITP(Institute for Information \& Communications Technology Planning \& Evaluation)

\bibliography{custom}

\appendix



\end{document}